\documentclass[letterpaper,10pt,conference]{IEEEtran}
\IEEEoverridecommandlockouts
\usepackage{cite}
\usepackage{amsmath,amssymb,amsfonts}
\usepackage[ruled,noend,linesnumbered]{algorithm2e}
\usepackage{booktabs}
\usepackage{tabularx}
\usepackage[breaklinks]{hyperref}
\makeatletter
\newcommand{\removelatexerror}{\let\@latex@error\@gobble}
\makeatother
\usepackage{graphicx}
\usepackage{textcomp}
\usepackage{xcolor}
\usepackage{subfigure}
\usepackage{hyperref}
\usepackage{cleveref}
\usepackage[all]{hypcap}
\usepackage{xcolor}
\hypersetup{
    colorlinks,
    linkcolor={red!50!black},
    citecolor={blue!50!black},
    urlcolor={blue!80!black}
}

\usepackage[colorinlistoftodos]{todonotes}
\usepackage{xcolor}
\usepackage[most]{tcolorbox}
\usepackage{todonotes}
\usepackage{color,soul}
\usepackage{tikz}
\usepackage[switch]{lineno}

\usepackage{scalerel}
\usetikzlibrary{svg.path}

\definecolor{orcidlogocol}{HTML}{A6CE39}
\tikzset{
  orcidlogo/.pic={
    \fill[orcidlogocol] svg{M256,128c0,70.7-57.3,128-128,128C57.3,256,0,198.7,0,128C0,57.3,57.3,0,128,0C198.7,0,256,57.3,256,128z};
    \fill[white] svg{M86.3,186.2H70.9V79.1h15.4v48.4V186.2z}
                 svg{M108.9,79.1h41.6c39.6,0,57,28.3,57,53.6c0,27.5-21.5,53.6-56.8,53.6h-41.8V79.1z M124.3,172.4h24.5c34.9,0,42.9-26.5,42.9-39.7c0-21.5-13.7-39.7-43.7-39.7h-23.7V172.4z}
                 svg{M88.7,56.8c0,5.5-4.5,10.1-10.1,10.1c-5.6,0-10.1-4.6-10.1-10.1c0-5.6,4.5-10.1,10.1-10.1C84.2,46.7,88.7,51.3,88.7,56.8z};
  }
}

\newcommand\orcidicon[1]{\href{https://orcid.org/#1}{\mbox{\scalerel*{
\begin{tikzpicture}[yscale=-1,transform shape]
\pic{orcidlogo};
\end{tikzpicture}
}{|}}}}

\begin{document}
\title{High Performance Im2win and Direct Convolutions using Three Tensor Layouts on SIMD Architectures}

\author{
\IEEEauthorblockN{
    Xiang Fu\IEEEauthorrefmark{1}
    Xinpeng Zhang\IEEEauthorrefmark{1}
    Jixiang Ma\IEEEauthorrefmark{1}
    Peng Zhao\IEEEauthorrefmark{2}
    Shuai Lu\IEEEauthorrefmark{1}
    Xu T. Liu\IEEEauthorrefmark{3}\IEEEauthorrefmark{4}\orcidicon{0000-0003-3980-9803}\thanks{This work is not related to the author's position at Amazon.}
    }
    \IEEEauthorblockA{
    \IEEEauthorrefmark{1}Nanchang Hangkong University, China
    \IEEEauthorrefmark{2}Microsoft, China
    \IEEEauthorrefmark{3}University of Washington, USA
    \IEEEauthorrefmark{4}Amazon Web Services, USA
    }
}

\maketitle

\begin{abstract}
Convolution is the core component within deep neural networks and it is computationally intensive and time consuming. Tensor data layouts significantly impact convolution operations in terms of memory access and computational efficiency. Yet, there is still a lack of comprehensive performance characterization on data layouts on SIMD architectures concerning convolution methods. This paper proposes three novel data layouts for im2win convolution: NHWC, CHWN, and CHWN8, and introduces a set of general optimization techniques for both direct and im2win convolutions. 
We compare the optimized im2win convolution with the direct convolution and PyTorch's im2col-based convolution across the aforementioned layouts on SIMD machines. The experiments demonstrated that the im2win convolution with the new NHWC layout achieved up to 355\% performance speedup over NCHW layout. 
Our optimizations also significantly improve the performance of both im2win and direct convolutions. 
Our optimized im2win and direct convolutions achieved up to 95\% and 94\% of machine's theoretical peak performance, respectively.
\end{abstract}


\section{Introduction}

Convolution is the essential component of deep neural networks for computer vision tasks such as feature exaction from large-scale image data\cite{cui_convolutional_2018}. It not only comprises 50\%-90\% of computational operations including convolutional, pooling, ReLU, and fully-connected layers\cite{shufflenet}, but also consumes more than 90\% of the total execution time of many popular neural networks\cite{efficient_processing_of_DNN,yan2024hierarchical,202406.1304}. Hence, optimizing convolution operations is crucial for enhancing the performance of neural networks.

Convolution methods can be classified into three main categories based on how they transform the input tensor: direct, im2col-based, and im2win. Direct convolution performs convolution operations directly on the tensor without changing its format\cite{direct_conv_on_SIMD}. This approach avoids extra memory consumption compared to im2col-based and im2win convolutions, but it suffers from nonconsecutive memory access.
Im2col-based convolution transforms the convolution into general matrix-matrix multiplications (GEMM)\cite{im2col_gpu_2006, p-im2col}, leveraging optimized Basic Linear Algebra Subprograms (BLAS)\cite{blas} for excellent performance. It has regular memory access but a significant extra memory footprint, which greatly limits its applicability on memory-constrained devices. 
Previously, we propose a memory-efficient convolution called image-to-window (im2win), which reorganizes the input tensor into a row of dot product windows and flattens the unique elements of these windows into a row that correspond to the convolution operation's receptive fields\cite{im2win_hpec_2022, lu_im2win_2023}. It provides sequential memory access and data reuse, and thus greatly reduces memory overhead. 

A tensor memory/data layout (referred as layout henceforth) refers to how the data of a tensor is physically arranged in memory. There are commonly three layouts for tensors: NCHW, NHWC and CHWN, where N is the batch size, C is the number of channels, H is the image height, and W is the image width. It significantly impacts convolution operations in terms of memory access, computational efficiency and compatibility with deep learning frameworks\cite{pytorch_nips_2019, tensorflow2015-whitepaper, cudnn_arxiv_2014}.

\begin{figure}[!ht]
\centering
\includegraphics[
  width=.5\textwidth,
  keepaspectratio,]{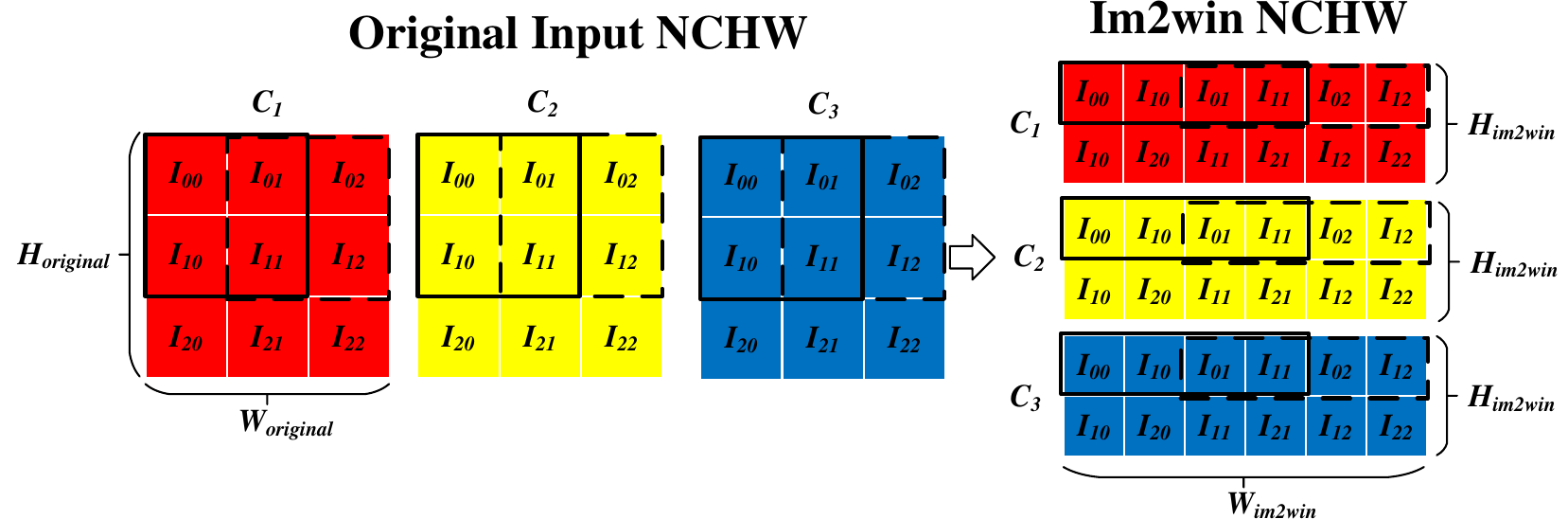}
\caption{\small The original input tensor ($N=1,H_{original}=W_{original}=C_{original}=3$) and its corresponding im2win tensor ($N=1,C_{im2win}=3,H_{im2win}=2,W_{im2win}=6$) in the NCHW layout}
\label{fig:nchw_layout}
\end{figure}

Tensors are stored as one-dimensional arrays in memory, although logically represented as four-dimensional arrays. An input tensor and an im2win tensor in a NCHW layout are illustrated in \Cref{fig:nchw_layout}. Different colors represent different channels. In the NCHW layout, the elements in the width dimension are contiguous in memory by prioritizing the width dimension first, followed by height, channel and batch. 
Assuming a stride of 1 and a filter tensor in the NCHW layout with dimensions of 1x3x2x2, the elements in the solid-lined boxes are used to compute the first output element in the output tensor, while the elements in the dashed-lined boxes are for the next. 
Further details will be provided in \Cref{sec:im2win}.

In general, the memory efficiency and performance implications of various tensor layouts with different convolution algorithms on single-input, multiple-data (SIMD) architectures have received limited attention.
The previous im2win works optimize the NCHW layout on CPU and GPU but have not tried the NHWC and CHWN layouts\cite{im2win_hpec_2022, lu_im2win_2023}.
Li et al. reveal the performance impact of the NCHW layout with the im2col-based convolution and the CHWN layout with the direct convolution in different CNN layers, and propose a fast multi-dimension layout transformation algorithm on GPU\cite{li_optimizing_2016}. 
Li et al. optimize the cache utilization of the NCHW and NHWC layouts in DNN training along with DNN pruning on CPU\cite{li_optimizing_2022}.
To date, no performance characterization have been conducted on NCHW, NHWC and CHWN layouts on SIMD architectures with the three convolution methods. 
 
To address aforementioned gaps, we propose three new layouts for the im2win convolution: NHWC, CHWN, and CHWN8,
and propose a set of optimization techniques based on the roofline model for both direct and im2win convolutions.
We compare the optimized im2win convolution with the optimized direct and PyTorch's im2col-based convolutions with the above layouts on SIMD machines.
Our experiments demonstrate that the new NHWC layout on the im2win convolution achieves 11\% to 355\% performance speedup compared with the NCHW layout. Our proposed optimizations have been empirically validated to enhance the performance of both im2win and direct convolutions. 
Our optimized im2win convolution and our optimized direct convolution achieve up to 95\% and 94\% of the theoretical peak performance of the machine, respectively.
As the main contributions, this work

1) proposes three novel layouts for im2win convolution.

2) comes up with a set of general optimization techniques that not only can be applied to im2win convolution and but also to direct convolution on different layouts.

3) compares the optimized im2win convolution with the im2col-based and the optimized direct convolutions using four tensor layouts on SIMD machines.

\section{Preliminary and Related Works}\label{sec:preliminary_and_related}

\subsection{Notation}
The input of a convolution operation includes an input tensor ($\mathcal{I}$), a filter tensor ($\mathcal{F}$), and an output tensor ($\mathcal{O}$). In these tensors,
$N_{i}$ is the batch size, $s$ is the stride size, $C_{i}$ and $C_{o}$ are the number of input and output channels (also known as feature maps), $H_{i/f/o}$ and $W_{i/f/o}$ denote the height and width of a feature map.

\subsection{Tensor Layouts: NCHW, NHWC, CHWN}

In the NCHW layout, the input ($\mathcal{I}$), the filter ($\mathcal{F}$), and the output tensors ($\mathcal{O}$) are expressed as $\mathcal{I}[N_{i}][C_{i}][H_{i}][W_{i}]$, $\mathcal{F}[C_{o}][C_{i}][H_{f}][W_{f}]$, and $\mathcal{O}[N_{i}][C_{o}][H_{o}][W_{o}]$, respectively. The convolution is defined as:
\begin{equation}
\begin{array}{r}
\begin{aligned}
\mathcal{O}_{(i, j, m, n)}=\sum_{j=1}^{C_{i}} \sum_{m=1}^{H_{f}} \sum_{n=1}^{W_{f}}\left(\mathcal{I}_{(i, j, m \times s+u, m \times s+v)}\right. \\
\left.\times \mathcal{F}_{(j, r, u, v)}\right),
\end{aligned}
\end{array}
\end{equation}

In the NHWC layout, the input ($\mathcal{I}$), the filter ($\mathcal{F}$), and the output tensors ($\mathcal{O}$) are expressed as $\mathcal{I}[N_{i}][H_{i}][W_{i}][C_{i}]$, $\mathcal{F}[C_{o}][H_{f}][W_{f}][C_{i}]$, and $\mathcal{O}[N_{i}][H_{o}][W_{o}][C_{o}]$, respectively. The convolution is defined as:
\begin{equation}
\begin{array}{r}
\begin{aligned}
\mathcal{O}_{(i, m, n, j)}=\sum_{j=1}^{C_{i}} \sum_{m=1}^{H_{f}} \sum_{n=1}^{W_{f}}\left(\mathcal{I}_{(i, m \times s+u, m \times s+v, j)}\right. \\
\left.\times \mathcal{F}_{(j, u, v, r)}\right),
\end{aligned}
\end{array}
\end{equation}

In the CHWN layout, the input ($\mathcal{I}$), the filter ($\mathcal{F}$) and the output tensors ($\mathcal{O}$) are expressed as $\mathcal{I}[C_{i}][H_{i}][W_{i}][N_{i}]$, $\mathcal{F}[C_{i}][H_{f}][W_{f}][C_{o}]$, and $\mathcal{O}[C_{o}][H_{o}][W_{o}][N_{i}]$ respectively. The convolution is defined as:
\begin{equation}
\begin{array}{r}
\begin{aligned}
\mathcal{O}_{(j, m, n, i)}=\sum_{j=1}^{C_{i}} \sum_{m=1}^{H_{f}} \sum_{n=1}^{W_{f}}\left(\mathcal{I}_{(j, m \times s+u, m \times s+v, i)}\right. \\
\left.\times \mathcal{F}_{(r, u, v, j)}\right),
\end{aligned}
\end{array}
\end{equation}
The definitions (1), (2) and (3) above are all subject to
\begin{equation}
\begin{aligned}
&j=1,2,..,C_{o}, m=1,2,..,H_{o}, n=1,2,..,W_{o}, \\
&i=1,2,..,N_{i}, u=1,2,..,H_{f}, v=1,2,..,W_{f},\\
&r=1,2,..,C_{i}. \nonumber
\end{aligned}
\end{equation}

\subsection{Convolution Algorithms and Related Works}

Direct convolution performs on the original $\mathcal{I}$ and $\mathcal{F}$ without any tensor transformation.
It has seven nested for loops and an AXPY operation in the innermost loop.
Based on the tensor layouts that direct convolution works with, the AXPY operation needs to read at different indices of $\mathcal{F}$ and $\mathcal{I}$, and writes at different indices of $\mathcal{O}$.

Direct convolution can compute with the original input tensors, so they usually adopt the NCHW layout of raw images. A study shows that, for convolution instances with large C, NHWC layout outperforms NCHW layout\cite{li_optimizing_2016}. Grouping a certain dimension of the input tensors by a fixed size can also enhance the performance of direct convolution, such as NC32HW32 layout\cite{ferrari_advancing_2023}.
Several works \cite{directconvolutions,direct_conv_on_SIMD} have shown that the performance of direct convolution can be greatly improved by designing specific layouts based on the loop ordering of the algorithm on SIMD architecture. 

The im2col-based convolution transforms a convolution operation into a GEMM operation. $\mathcal{I}[N_{i}][C_{i}][H_{i}][W_{i}]$ is processed in $N_i$ batches, each batch contains data $\mathcal{I'}[C_i][H_i][W_i]$ (that is, a single image). 
The im2col algorithm flattens the elements of each dot product window of  $\mathcal{I'}$ and copies them into a single row of a 2D matrix\cite{im2col_gpu_2006}.
In addition to the conventional im2col data transformation algorithm, 
the MEC algorithm compresses the matrix layout, while still enabling the utilization of high-performance BLAS algorithms to perform convolution operations\cite{mec}.

\section{High-performance Im2win and Direct Convolution using Three Tensor Layouts}\label{sec:im2win}

In this section, we first review three layouts in the context of the direct convolution. Then we present three new layouts for the im2win convolution, following by how to determine the loop ordering based on the layouts and a set of optimizations for the im2win and direct convolutions on SIMD architectures.

\begin{figure*}[!ht]
\centering
\includegraphics[
  width=\textwidth,
  keepaspectratio,]{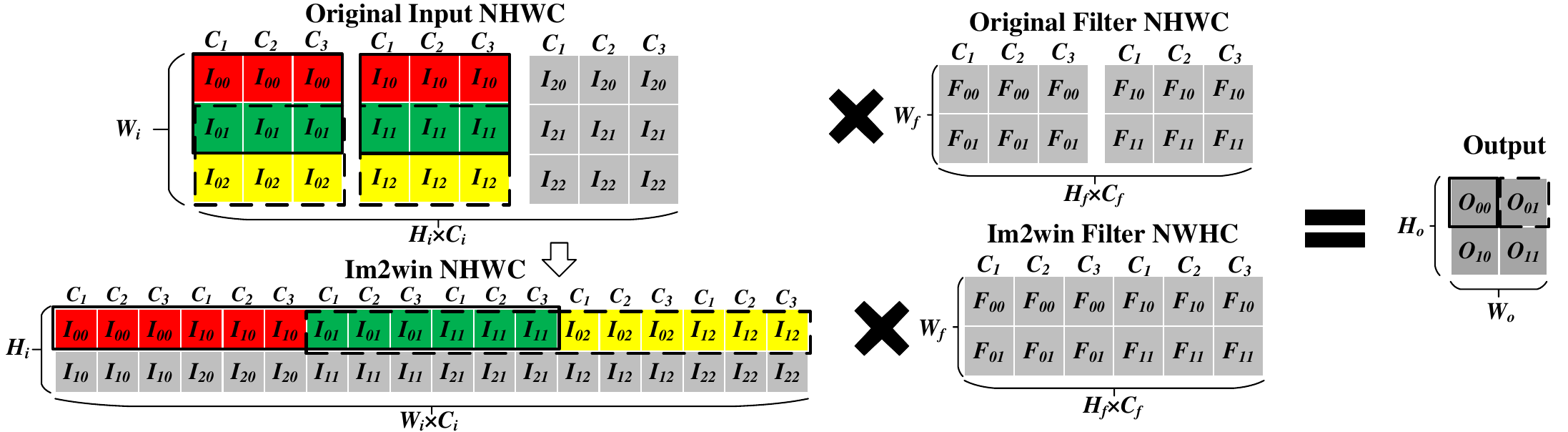}
\caption{\small The original input tensor ($N_{i}=1,H_{i}=W_{i}=C_{i}=3$) and its corresponding im2win tensor ($N_{i}=1,C_{i}=3,H_{i}=2,W_{i}=6$) in the NHWC layout, the filter tensor ($N_{f}=1,C_{f}=3,H_{f}=W_{f}=2$), $s=1$, the output tensor ($N_{o}=1,C_{o}=1,H_{o}=W_{o}=2$)}
\label{fig:nhwc_convolution}
\end{figure*}

\begin{figure}[!ht]
	\centering
	\includegraphics[width=.5\textwidth,
  keepaspectratio,]{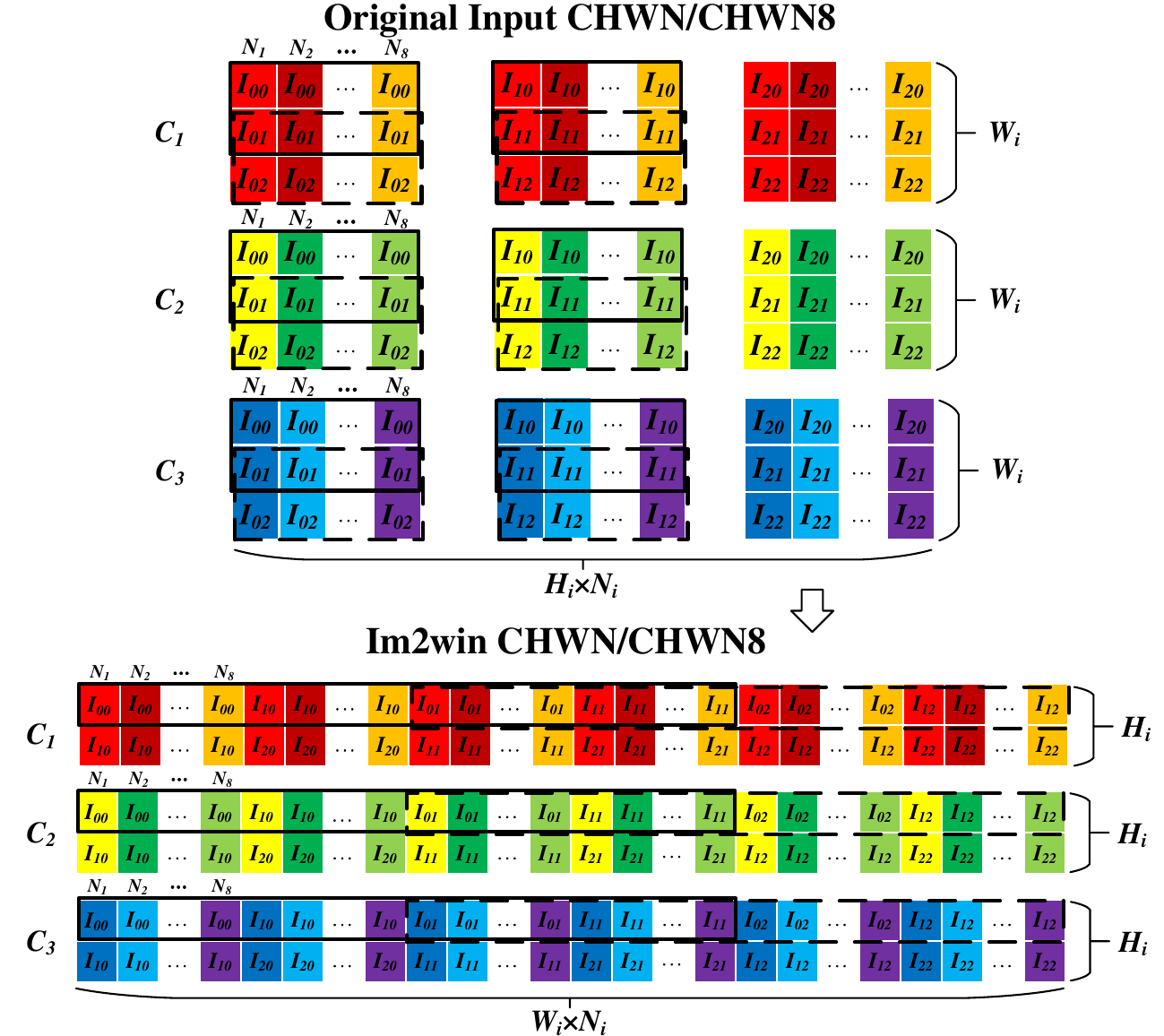}
	\caption{\small The original tensor ($N_{i}=8,H_{i}=W_{i}=C_{i}=3$) and its corresponding im2win tensor ($N_{i}=8,C_{i}=3,H_{i}=2,W_{i}=6$) in the CHWN/CHWN8 layout}
	\label{fig:chwn_layout}
\end{figure}

\subsection{Motivations for Different Tensor Layouts}\label{input three layouts}

An example of direct convolution on an original input tensor in the NHWC layout is illustrated in \Cref{fig:nhwc_convolution}. 
The NHWC layout prioritizes the storage of elements in the last logical dimension—$C_{i}$—followed by that of $W_{i}$, $H_{i}$, and $N_{i}$. 
In the NHWC layout, the elements with the same $N_{i}$, $H_{i}$, and $W_{i}$ but different $C_{i}$ are contiguous in memory, which have unit stride access.
Assuming a stride of 1,
the red and green elements (outlined by solid lines, representing the convolutional windows to compute a single output element) in the input tensor are multiplied with the corresponding elements in the filter tensor, and the results are summed up (i.e., an AXPY) to obtain $O_{00}$. 
Next, the green and yellow elements (outlined by dashed lines) are used to calculate $O_{01}$. This process continues, with the convolutional window moving by the stride length in $H_{i}$ or $W_{i}$, until all elements of the output tensor are computed.

Recall in \Cref{fig:nchw_layout}, the NCHW layout prioritizes the storage of elements in the last logical dimension—$W_{i}$—followed by that of $H_{i}$, $C_{i}$, and $N_{i}$.
Note that all $C_{i}$ but not all $W_{i}$ are used during the AXPY operation to obtain one output element. NCHW has non-unit stride access during the tensor convolution. However, the non-unit stride access may not be harmful, depending on the effects of caching and the access patterns used (determined by the loop ordering).

As shown in \Cref{fig:chwn_layout}, the CHWN layout stores elements by prioritizing $N_{i}$, followed by $W_{i}$, $H_{i}$, and $C_{i}$ in memory. 
Previous research on GPU recommends to use N as the lowest dimension for coalesced memory access and data reuse in registers\cite{li_optimizing_2016}. It has been observed that the performance is sensitive to the value of N.
The elements within the solid-lined boxes are used to compute the first eight output elements, while the elements within the dashed-lined box are used for the next eight output elements. 
This facilitates the use of vector registers for vectorization.
Assuming $s=1$, the convolutional window moves by one element in both the $H_i$ and $W_i$ dimensions until all output elements are computed.

{
\begin{algorithm}[t]
\small
\SetAlgoNoLine
    \caption{\small Im2win Tensor Transformation Algorithm}
    \label{algorithm:im2winTransform}
    \KwIn{input tensor $\mathcal{I}$ in the NHWC layout, filter tensor $\mathcal{F}$, Stride $s$}
    \KwOut{Im2win tensor $\mathcal{\hat I}$ in the NHWC layout}
    $H_o = (H_i-H_f)/s+1$ \label{algorithm:im2winTransform:line:calculate_Ho}\\
    \For{$i=1$ \bf{to} $N_{i}$ }{\label{algorithm:im2winTransform:line:copy_Ni}
         \For{$m=1$ \bf{to} $H_{o}$}{\label{algorithm:im2winTransform:line:copy_Ho}
            \For{$k=1$ \bf{to} $W_{i}$}{\label{algorithm:im2winTransform:line:copy_Wi}
                \For{$u=1$ \bf{to} $H_{f}$}{\label{algorithm:im2winTransform:line:copy_Hf}
                    \For{$r=1$ \bf{to} $C_{i}$}{\label{algorithm:im2winTransform:line:copy_Ci}
                $\mathcal{\hat I}[i][m][ u + k \times H_f][r]$ = $\mathcal{I}[i][m \times s +u][k][r]$
                    }}}}}\label{algorithm:im2winTransform:line:copy}
\end{algorithm} 
}

{
\begin{algorithm}[t]
\small
\SetAlgoNoLine
    \caption{\small Naive Im2win Convolution}
    \label{algorithm:im2winConvolution}
    \KwIn{input tensor $\mathcal{I}$ in the NHWC layout, filter tensor $\mathcal{F}$ in the NHWC layout, Stride $s$}
    \KwOut{output tensor $\mathcal{O}$ in the NHWC layout}
    $\mathcal{\hat I}$ = \textbf{Function} \textsc{im2win}($\mathcal{I},\mathcal{F},s$)
\SetKwProg{func}{Function}{}{}\\
transform $\mathcal{F}$ in NHWC to $\mathcal{\hat F}$ in NWHC \\
    \For{$i=1$ \bf{to} $N_{i}$}{\label{algorithm:im2winConvolution:line:copy_No}
        \For{$m=1$ \bf{to} $H_{o}$}{\label{algorithm:im2winConvolution:line:copy_Ho}        
            \For{$j=1$ \bf{to} $C_{o}$} 
            {\label{algorithm:im2winConvolution:line:copy_Co}
                \For{$n=1$ \bf{to} $W_{o}$}{\label{algorithm:im2winConvolution:line:copy_Wo}
                    \For{$v=1$ \bf{to} $W_{f}$}{\label{algorithm:im2winConvolution:line:copy_Wf}
                        \For{$u=1$ \bf{to} $H_{f}$}{\label{algorithm:im2winConvolution:line:copy_hf}
                            \For{$r =1$ \bf{to} $C_{i}$}{\label{algorithm:im2winConvolution:line:copy_cf}
                                $\mathcal{O}[i][m][n][j]$ += $\mathcal{\hat I}[i][m][n \times s \times W_{f} + v \times W_{f} + u][r]$ $\times$ $\mathcal{\hat F}[j][v][u][r]$}}}}}}}\label{algorithm:im2winConvolution:line:axpy}
\end{algorithm}
}

\subsection{Im2win Tensor Transformation on Three Tensor Layouts}\label{im2win three layouts}

In this subsection, we present the im2win tensor transformation on three tensor layouts. The im2win tensor transformation process for the NWHC layout is shown as \Cref{algorithm:im2winTransform}. For other layouts, slight modifications need to be made on it. The im2win transformation flattens the elements of a convolutional window, storing them contiguously in memory and prioritizing them in the $C_i$ dimension, as outlined in \Cref{algorithm:im2winTransform} from \Cref{algorithm:im2winTransform:line:copy_Wi} to \Cref{algorithm:im2winTransform:line:copy}. 
An example of transforming the original input tensor into an im2win tensor in the NHWC layout is shown in \Cref{fig:nhwc_convolution}. It is important to note that the green elements can be reused between two adjacent windows, which eliminates the need for repetitive storage in memory like the im2col transformation\cite{im2col_gpu_2006}.
In the output im2win tensor of \Cref{algorithm:im2winTransform}, the elements involved in each convolution operation are contiguous and more compact in memory, 
improving spatial locality,  cache and SIMD efficiency.

An example of the im2win convolution using the NHWC layout is shown in \Cref{fig:nhwc_convolution}. A naive im2win convolution method is shown as \Cref{algorithm:im2winConvolution}. It is similar to the direct convolution, which involves seven nested loops and an AXPY in the innermost loop, 
except prior to that, the im2win convolution performs a tensor transformation upon $\mathcal{I}$ to get $\mathcal{\hat I}$, and $\mathcal{F}$ in the NHWC layout is transformed into a NWHC layout to match $\mathcal{\hat I}$.
All output elements can be computed through the outer four loops of \Cref{algorithm:im2winConvolution} from \Cref{algorithm:im2winConvolution:line:copy_No} to \Cref{algorithm:im2winConvolution:line:copy_Wo}.

An im2win tensor in CHWN/CHWN8 layout is shown in \Cref{fig:chwn_layout}. The characteristics of $C_{i}$, $H_{i}$, and $W_{i}$ in this layout are similar to NCHW, but the CHWN layout prioritizes storage in $N_{i}$. 
The CHWN layout's efficiency is constrained by 256-bit vector registers, which process only 8 outputs at once. When $N_i > 8$, this leads to low cache utilization, as the cache holds unnecessary data for these 8 output calculations.
Motivated by the approach of dividing $C_{i}$ into blocks in the direct convolution \cite{directconvolutions}, we propose a new CHWN8 layout.
CHWN8 lays 8 $N_{i}$ in the innermost layer and remaining $N_{i}$ in the outermost layer, which takes full advantage of the vector registers without sacrificing the cache utilization.
$N_{i}$ can be set to a multiple of 8 (with padding if necessary), 
CHWN8 layout may provide better data continuity than the NHWC layout when $C_{i}$ is relatively small.

{
\begin{algorithm}[t]
\small
\SetAlgoNoLine
\caption{\small High Performance Im2win Convolution}
\label{algorithm:High_performance_im2win}
\KwIn{input tensor $\mathcal{I}$ in NHWC layout, filter tensor $\mathcal{F}$ in NHWC layout, Stride $s$}
\KwOut{output tensor $\mathcal{O}$ in NHWC layout}
$\mathcal{\hat I}$ = \textbf{Function} \textsc{im2win}($\mathcal{I},\mathcal{F},s$)\\
transform $\mathcal{F}$ in NHWC to $\mathcal{\hat F}$ in NWHC \\
$H_o = (H_i-H_f)/s+1$
\label{algorithm:High_performance_im2win:line:calculate_Ho}\\
\SetKwProg{func}{Function}{}{}
\For{$im=1$ \bf{to} $N_{i} \times H_{o}$ \bf{in parallel}}{
{\label{algorithm:High_performance_im2win:line:outer_loop_begin}}
$i = im / H_o$\label{algorithm:High_performance_im2win:line:calculate_i}, $m = im \% H_o$ \label{algorithm:High_performance_im2win:line:calculate_m}\\
    \For{$j=1$ \bf{to} $C_{o}$}{\label{algorithm:High_performance_im2win:line:calculate_Co}
        \For{$n=1$ \bf{to} $W_{o} / W_{o,b}$}{
        {\label{algorithm:High_performance_im2win:line:outer_loop_end}} 
         \sc{dot\_product}($i,j,m,n,W_{o,b},s$)
}}}

\func{\textsc{dot\_product}($i,j,m,n,W_{o,b},s$):}{
$ymm_1$ = ... = $ymm_{W_{o,b}}$ = 0\\
\For{$r=1$ \bf{to} $C_i$}{
{\label{algorithm:High_performance_im2win:line:convolution_window_Cf}}
    \For{$v=1$ \bf{to} $W_{f}/N_{vec}$}{
    {\label{algorithm:High_performance_im2win:line:convolution_window_Wf}}
         \For{$u=1$ \bf{to} $H_{f}$}{
        {\label{algorithm:High_performance_im2win:line:convolution_window_Hf}}
        FMA($\mathcal{\hat{I}}[i][r][u + m][v \times N_{vec}]$, $\mathcal{\hat{F}}[j][r][u][v]$, $ymm_1$)\\
      {\label{algorithm:High_performance_im2win:line:FMA}}
        ...\\
        FMA($\mathcal{\hat{I}}[i][r][u + m][v \times N_{vec} + s \times (W_{o,b}-1)]$, $\mathcal{\hat{F}}[j][r][u][v]$, $ymm_{W_{o,b}}$)
       
}}}}

\end{algorithm}
}

\subsection{Loop Reordering}

In this subsection, we reorder the loops for the im2win and direct convolutions based on different tensor layouts.
Ideally, we arrange the inner loops to access data closer in memory to enjoy the unit stride access as long as possible.
In both direct and im2win convolutions, for NCHW, CHWN, and CHWN8 layouts, as they have the CHW memory access pattern, we use the width of the convolution window as the innermost loop, followed by the height, and the channel ($W_f$, $H_f$, and $C_i$). 
Conversely, for the NHWC layout, the innermost three loops iterate over the channel, the width, and the height ($C_i$, $W_f$, and $H_f$) of the convolution window. 
For the im2win convolution, since the im2win transformation flattens the elements of a convolutional window, we usually coalesce the height and the width layers ($W_f, H_f$) into a $W_f \times H_f$.

Next we determine the order of the outer four loops. It not only determines the order in which the elements of the output tensor are produced but also influences the order in which the element in the input tensor are accessed. 
The layout mainly affects the inner three levels of the loop order, for the outer four levels the loop order applies to all data layouts.
Recall in \Cref{fig:nhwc_convolution}, the green elements of the input tensor are shared between the convolution windows of adjacent output tensor elements. Therefore, we consider to position the width ($W_o$) of the traversed output tensor in the fourth layer of loops at \Cref{algorithm:im2winConvolution:line:copy_Wo} in \Cref{algorithm:im2winConvolution}.
Because the memory access of the input tensor is expensive in a convolution operation, we set the channel ($C_o$) as the third layer to reduce its access.
Since the batch of the output tensor corresponds to the batch of the input tensor, we place the batch ($N_i$) in the first layer of the loop. 
Finally we have the order of the outer four loops as NHCW in \Cref{algorithm:im2winConvolution}. 

\subsection{Optimizations for the Im2win and Direct Convolutions}\label{sec:optimizations}

In this subsection, we use the Roofline Model\cite{samuel_roofline_2009} to determine how to optimize the im2win and direct convolutions on SIMD systems. 
We propose a set of optimizations for both the im2win and direct convolutions and apply the optimizations to them (with a slight modification on the direct convolution). The optimizations are classified into two categories: reducing the memory bottleneck and increasing the arithmetic intensity of the convolution kernel. The former includes hoist, memory alignment, register and cache blocking. The latter includes
loop unrolling, vectorization and FMA instructions, loop coalescing, and parallelization strategies.
We have three level parallelization: Non-Uniform Memory Access (NUMA) level, thread level using OpenMP and instruction level using SIMD.

We present the optimized im2win convolution as \Cref{algorithm:High_performance_im2win}. 
Since each element of the output tensor in  each batch ($N_i$) can be computed independently, there is abundant parallelism available\cite{Anatomyofhigh} and this is NUMA-friendly.
Within each batch, the shared elements of adjacent windows can be maximized across the threads.
When the dimension we choose to parallelize is small and the number of CPU cores is large, this leads to a workload imbalance in many core architecture. The solution is to coalesce multiple dimensions/loops into one parallel loop to achieve better load balance. In practice, we find that coalescing two dimensions of the output tensor yields the best load balance. 
Hence, we apply parallel strategies and coalesce $N_i$ and $H_o$ in a parallel loop at \Cref{algorithm:High_performance_im2win:line:outer_loop_begin} in~\Cref{algorithm:High_performance_im2win}.

In \Cref{algorithm:High_performance_im2win}, we hoist three things: the indices of the elements in the 1D array of the tensor, the elements of the input tensor, and the entire filter tensor at \Cref{algorithm:High_performance_im2win:line:convolution_window_Wf}.
We use register and cache blocking\cite{Register_blocking_wolfe_1987, loop_tiling_wolfe_1989} at \Cref{algorithm:High_performance_im2win:line:outer_loop_end} to reduce cache misses because the neighboring convolutional windows share duplicate elements of the input tensor, which $W_{o,b}$ is the blocking size.
Next, we unroll the loop at \Cref{algorithm:High_performance_im2win:line:convolution_window_Wf} to take advantage of the spatial locality,
because the innermost loop accesses consecutive elements.
Since we are using FP32 and AVX2, we set the loop unrolling size $N_{vec}$=8. Finally, we vectorize the input tensor and filter tensor in units of eight to use the FMA instruction in AVX2\cite{im2win_hpec_2022} at \Cref{algorithm:High_performance_im2win:line:FMA}. 

A cache-line is 512 bits on contemporary x86\_64 architectures, which is the minimum data quantity that can be fetched from memory to cache. Without memory alignment of the tensor data structure, the CPU needs to issue two memory requests to access an element, because the element may be in the middle of the cache-line. This hurts the performance greatly because not only the CPU has to wait for memory access but also more caches are used. This leads to higher cache miss and lower cache utilization. We store the elements of the tensor in a memory-aligned way using posix\_memalign in C during the memory allocation.

{
\begin{table}[ht]
\caption{\small Twelve convolution layers of the DNN benchmarks.}
\label{table:benchmarks}
\begin{center}
\resizebox{\linewidth}{!}{
\begin{tabular}{cccc}
\toprule
\textbf{NAME}&\textbf{INPUT}&\textbf{FILTER, STRIDE}&\textbf{OUTPUT} \\
& $C_{i} \times H_{i} \times W_{i}$ & $C_{o} \times H_{f} \times W_{f}, s_{h}/s_{w}$&$C_{o} \times H_{o} \times W_{o}$\\
\midrule
$\textbf{conv1}$ & $3\times227\times227$& $96\times11\times11, 4$&$96\times55\times55$\\
$\textbf{conv2}$ & $3\times231\times231$& $96\times11\times11, 4$&$96\times56\times56$\\
$\textbf{conv3}$ & $3\times227\times227$& $64\times7\times7, 2$&$64\times111\times111$\\
$\textbf{conv4}$ & $64\times224\times224$& $64\times7\times7, 2$&$64\times109\times109$\\
$\textbf{conv5}$ & $96\times24\times24$& $256\times5\times5, 1$&$256\times20\times20$\\
$\textbf{conv6}$ & $256\times12\times12$& $512\times3\times3, 1$&$512\times10\times10$\\
$\textbf{conv7}$ & $3\times224\times224$& $64\times3\times3, 1$&$64\times222\times222$\\
$\textbf{conv8}$ & $64\times112\times112$& $128\times3\times3, 1$&$128\times110\times110$\\
$\textbf{conv9}$ & $64\times56\times56$& $64\times3\times3, 1$&$64\times54\times54$\\
$\textbf{conv10}$ & $128\times28\times28$& $128\times3\times3, 1$&$128\times26\times26$\\
$\textbf{conv11}$ & $256\times14\times14$& $256\times3\times3, 1$&$256\times12\times12$\\
$\textbf{conv12}$ & $512\times7\times7$& $512\times3\times3, 1$&$512\times5\times5$\\
\bottomrule
\end{tabular}}
\end{center}
\end{table}
}

{
\begin{figure*}[ht]
	\centering
	\includegraphics[width=.95\textwidth]{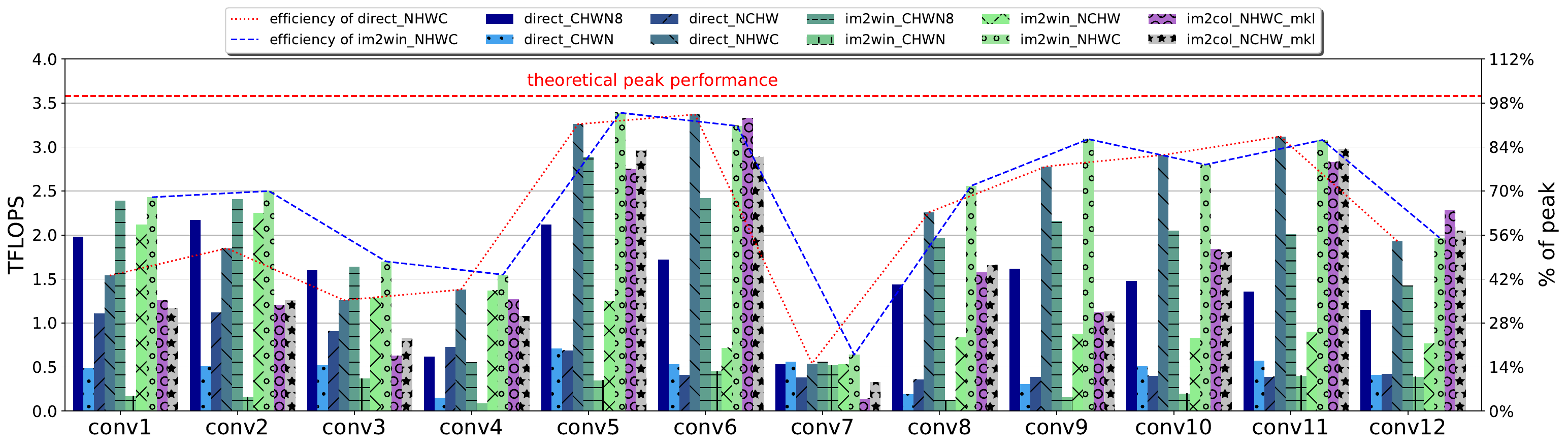}
	\caption{\small Performance results in TFLOPS of the direct convolution, the im2win convolution and the im2col-based convolution using different layouts. Note that the theoretical peak performance of the server is 3584 GFLOPS.}\label{fig:overall_flops}
\end{figure*}

\begin{figure*}[ht]
	\centering
	\includegraphics[width=.95\textwidth]{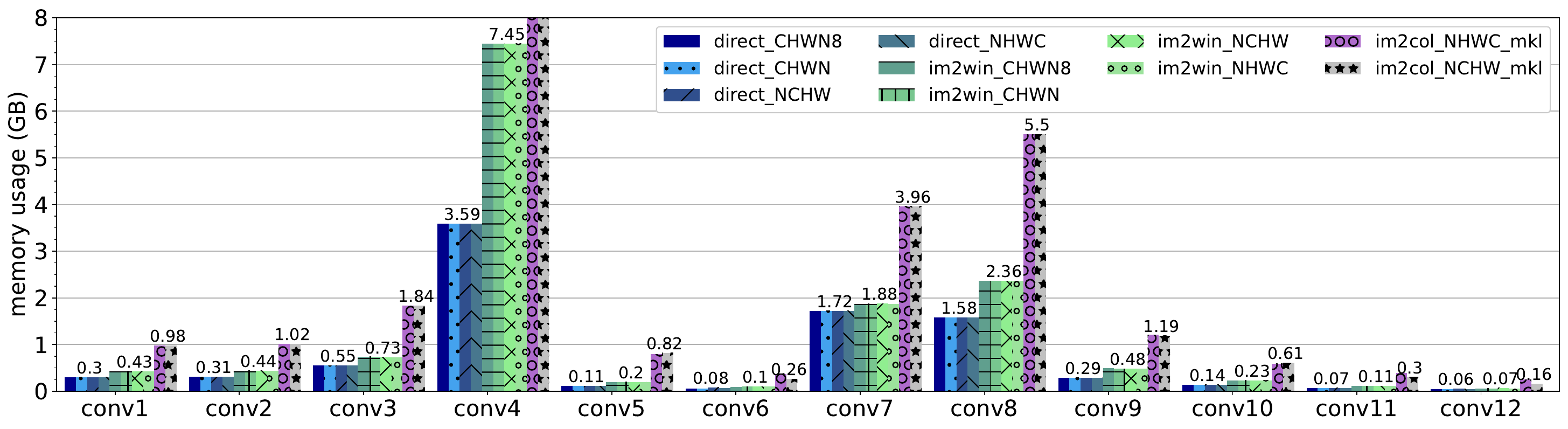}
	\caption{\small Memory usage of the direct, the im2win and the im2col-based convolutions using different tensor layouts. Note that in conv4, the im2col-based convolutions with the NWHC and NCHW layouts use 21GB of memory.}
	\label{fig:overall_memory}
\end{figure*}
}

\section{Experiments}\label{sec:experiments}
\subsection{Experimental Setup}
\noindent\textbf{Architectures.} We use a server with two Intel\textsuperscript{\small{\textregistered}} Xeon\textsuperscript{\small{\textregistered}} Gold 6330 CPUs and 251 GB RAM. Each CPU has 28 physical cores, running at 2.0 GHz, with 48 KB L1d cache, 32 KB L1i cache, 1.28MB L2 cache and 43 MB L3 cache.

\noindent\textbf{Benchmarks} We aim to cover the majority of the convolutional layers in commonly used DNNs. 
Hence for our experimental evaluation, we select an state-of-the-art DNN benchmark\cite{mec} shown in~\Cref{table:benchmarks}, which includes twelve unique convolution layers, conv1-conv12.

\noindent\textbf{Software} We compare the direct, im2win, and im2col-based convolutions with four tensor layouts: NHWC, NCHW, CHWN, CHWN8. We intend to compare with the state-of-the-art implementation of direct convolution and the layout proposed in\cite{directconvolutions}, but their implementation is not open-sourced. 
We use the im2col-based convolution in PyTorch 2.1\cite{pytorchlib} with MKL\cite{intel_mkl}. Note that PyTorch only supports the NHWC and NCHW layouts.
Our code is compiled with GCC 9.5.0 compiler and \texttt{-O3 -mavx2 -mfma -fopenmp -march=native} compilation flags. We use OpenMP 4.0 for parallelization with \textit{guided} scheduling.

\subsection{Performance of Different Convolution Algorithms}

We run each algorithm 50 times on each benchmark with $N_i$=128 and report the best runtime. 
\Cref{fig:overall_flops} shows the performance results in TFLOPS and~\Cref{fig:overall_memory} shows the memory usage of our high-performance direct convolution and high-performance im2win-based convolution, and the im2col-based convolution using MKL in PyTorch. In~\Cref{fig:overall_flops}, the left y-axis shows the performance in TFLOPS, and the right y-axis shows the performance of the machine peak.
Overall, our optimized im2win convolution achieves eight best TFLOPS out of twelve benchmarks; our optimized direct convolution achieves three best TFLOPS out of twelve benchmarks and the im2win convolution achieves close performance on these three; the im2col-based convolution in PyTorch achieves one out of twelve best TFLOPS on conv12. 
All twelve best TFLOPS are all yielded from the NHWC layout across these three methods.
Our proposed optimization techniques are proven effective on both im2win and direct convolutions. Our im2win convolution achieves 95\% and 91\% of the theoretical peak performance of the architecture on conv5 and conv6 respectively. Our direct convolution achieves 91\% and 94\% of the theoretical peak performance on conv5 and conv6 respectively. 

With the NHWC layout and performance normalization, 
excluding conv6 and conv12, our im2win convolution achieves between 1.1$\times$ and 4.6$\times$ performance speedup, and our direct convolution achieves between 1.1$\times$ and 3.8$\times$ performance speedup against the im2col-based convolution. All the best performance of the im2win convolution on twelve benchmarks is achieved using the NHWC layout on CPU. Our im2win convolution using the NHWC layout outperforms the NCHW layout by at least 11\% and up to 355\% across all benchmarks, showcasing significant efficiency gains in various scenarios.
For the direct convolution, nine out of twelve best TFLOPS are achieved using the NHWC layout on CPU, the rest three are achieved using the CHWN8 layout.

With the NCHW layout,
our im2win-based convolution achieves between 1.4$\times$ and 2.4$\times$ performance speedup on all benchmarks, and PyTorch's im2col-based convolution achieves between 1.1$\times$ and 7.5$\times$ performance speedup (excluding conv7) against the direct convolution. The direct convolution performs poorly on the NCHW layout. 

With the CHWN/CHWN8 layout, im2win\_CHWN8 outperforms im2win\_CHWN on all benchmarks from 3.7$\times$ to 16$\times$; direct\_CHWN8 outperforms 2.3$\times$ and 8$\times$ over direct\_CHWN except in conv7. 
This shows our proposed CHWN8 layout overwhelmingly beats the CHWN layout on these two methods.
The direct convolution with the CHWN8 layout 
 performs better than the CHNW, NCHW and NHWC layouts when $C_i$ is small ($C_i=3$ for conv1, conv2, conv3). Similar results have also been observed in the previous GPU work\cite{li_optimizing_2016}.

In~\Cref{fig:overall_memory}, using the same tensor layout, the memory usage of each convolution is the same hence we only annotate one number per method. In all benchmarks, the direct convolution uses the least memory, while the im2col-based convolution consumes the most memory.
On average, the im2col-based convolution has 3.9$\times$ more memory usage than direct convolution, and im2win-based convolution has 1.5$\times$ more memory usage than direct convolution. The im2win-based convolution uses on average 39\% of the memory of the im2col-based method, and in conv5, it uses only 24\% of the memory of the im2col-based convolution.

\section{Conclusion}\label{sec:conclusion}

We proposed three new layouts for the im2win convolution: NHWC, CHWN and CHWN8, and a set of general optimization techniques for both direct and im2win convolutions on SIMD architectures. We applied these optimizations on the im2win and the direct convolutions, and compared with PyTorch's im2col-based convolution on the above layouts along with the NCHW layout. 
Our experiments demonstrated that im2win convolution using the new NHWC layout achieved 11\% to 355\% performance speedup compared to NCHW layout. 
The proposed optimizations were proven to boost the performance of the im2win convolution and direct convolution. 
Our optimized im2win and direct convolutions achieved up to 95\% and 94\% of the theoretical peak performance of the machine, respectively.

\section*{Acknowledgment}
This work was partially supported by the NSF OAC-2402542, and the National Natural Science Foundation of China (Grant No. 42261070).

\bibliographystyle{IEEEtran}
\bibliography{bibliography.bib}

{
\newpage

\appendix\label{sec:appendix}

We leave some information out from the main body of this paper and have them as part of an appendix. Because we consider they are not essential to the main argument but may be useful for readers who want to dive deeper into the topic. 

\subsection{Peak Performance}
We use the following formula to calculate the peak-gflops of the server:

\begin{multline} \label{eq:peak_gflops}
    \frac{peak_{flop}}{s} = (\#processors) \times (\#cores_{per\_processor}) \\
    \times (clock_{speed}[1/s]) \times (2 \times \#{FMA}_{units}) \times \frac{vector_{size}[bits]}{64}
\end{multline}
Based on~\Cref{eq:peak_gflops}, the peak GFLOPS of the server that we use in our experimental evaluation is 3584 GFLOPS.

\subsection{Batch Size Scaling on Different layouts}

We perform a strong scaling on the batch size from 32, 64, 128, 256 to 512 with different layouts using the direct convolution and the im2win convolution.
The performance results of the direct convolution with the CHWN, CHWN8, NCHW, and NHWC layouts are shown in \Cref{fig:direct_chwn_batch_scaling}, \Cref{fig:direct_chwn8_batch_scaling}, \Cref{fig:direct_nchw_batch_scaling}, and \Cref{fig:direct_nhwc_batch_scaling}, respectively.
The performance results of the im2win convolution with the CHWN, CHWN8, NCHW, and NHWC layouts are shown in \Cref{fig:im2win_chwn_batch_scaling}, \Cref{fig:im2win_chwn8_batch_scaling}, \Cref{fig:im2win_nchw_batch_scaling}, and \Cref{fig:im2win_nhwc_batch_scaling}, respectively.

From the figures, we can tell that the CHWN layout is most sensitive to the batch size among four layouts. The performance of the direct and im2win convolutions is the best on twelve benchmarks when the batch size is 32 (except for the direct convolution on conv12). Recall in \Cref{im2win three layouts}, the efficiency of the CHWN layout is constrained by the number of vector registers on the SIMD machines. When $N_i>8$, this leads to low cache utilization.

With our proposed CHWN8 layouts, the performance of the direct convolution and the im2win convolution exhibits similar patterns on twelve benchmarks, that is, when the channel sizes of the benchmarks are small ($C_i$=3 for conv1, conv2, conv3), the smaller the batch size is, the better performance the convolutions have on these benchmarks; when the channel sizes of the benchmarks (conv4-conv12) are large, the larger the batch size is, the better performance the convolution have on these benchmarks.

With the NWHC and NCHW layouts, both convolution methods show no obvious evidence that they are sensitive to the batch size across all benchmarks. The performance variance is contributed by the overall dimension of the input/filter tensors, and the loop coalescing which we coalesce the $N_i$ and $H_o$ dimensions into one parallel loop to achieve better load balance. 



\begin{figure*}[ht]
	\centering
	\includegraphics[width=.9\textwidth]{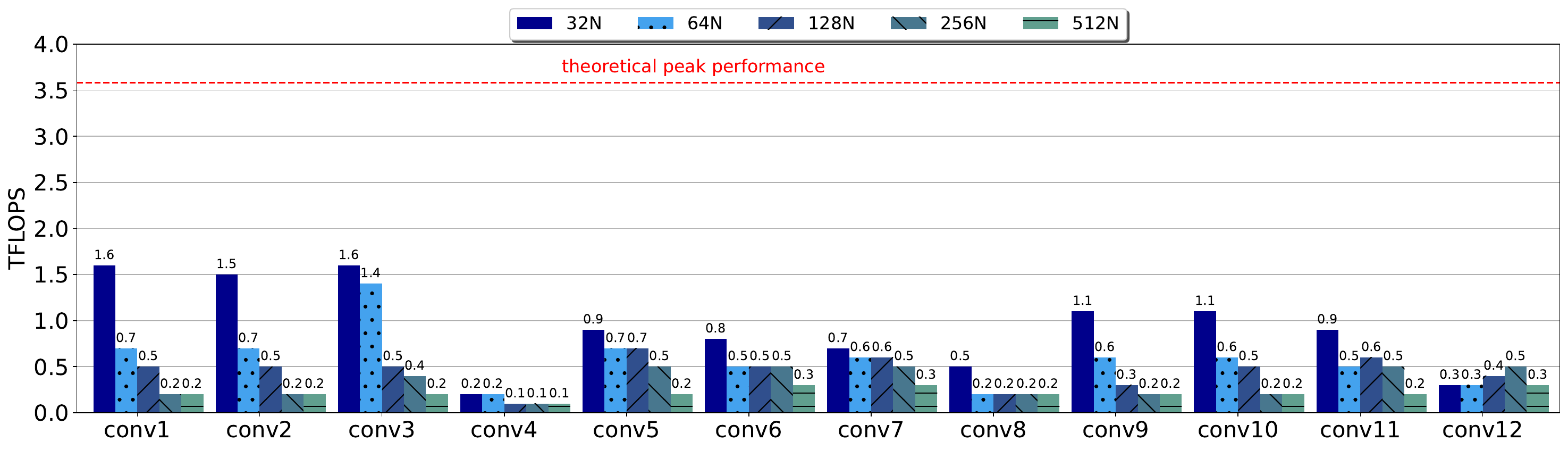}
	\caption{\small The performance of the direct convolution in different batch sizes with the CHWN layout}
	\label{fig:direct_chwn_batch_scaling}
\end{figure*}

\begin{figure*}[ht]
	\centering
	\includegraphics[width=.9\textwidth]{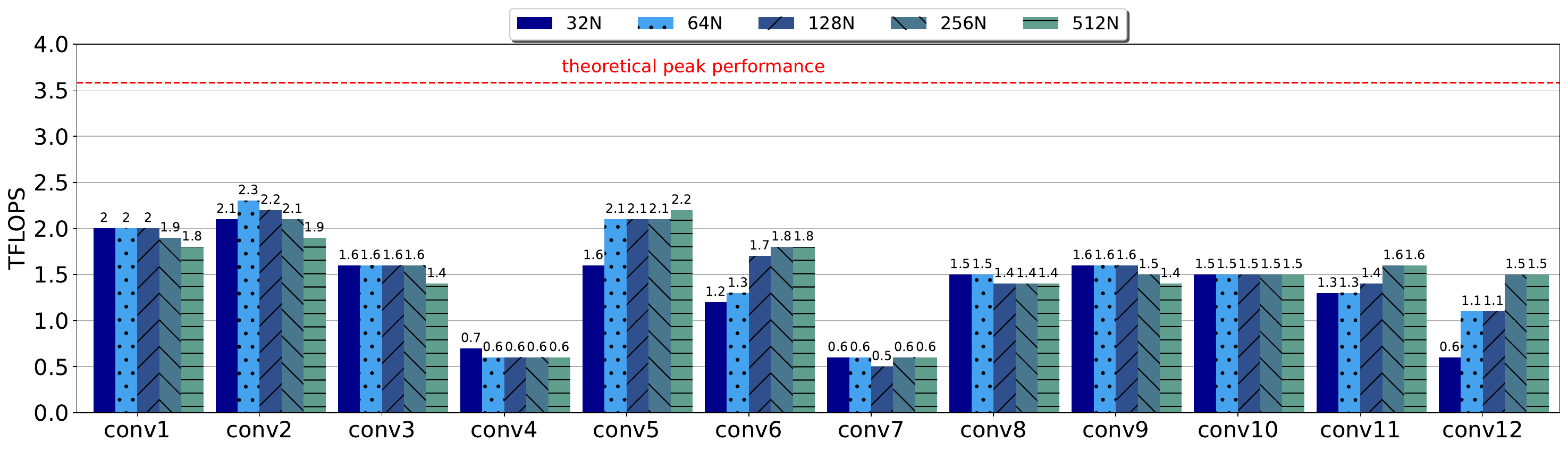}
	\caption{\small The performance of the direct convolution in different batch sizes with the CHWN8 layout}
	\label{fig:direct_chwn8_batch_scaling}
\end{figure*}

\begin{figure*}[ht]
	\centering
	\includegraphics[width=.9\textwidth]{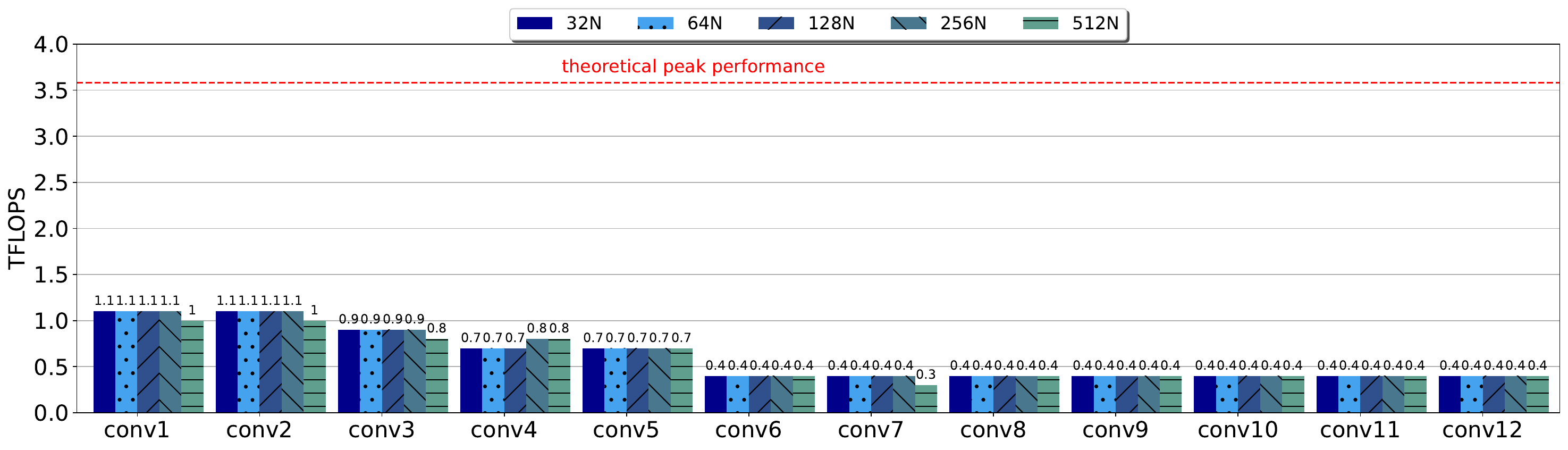}
	\caption{\small The performance of the direct convolution in different batch sizes with the NCHW layout}
	\label{fig:direct_nchw_batch_scaling}
\end{figure*}

\begin{figure*}[ht]
	\centering
	\includegraphics[width=.9\textwidth]{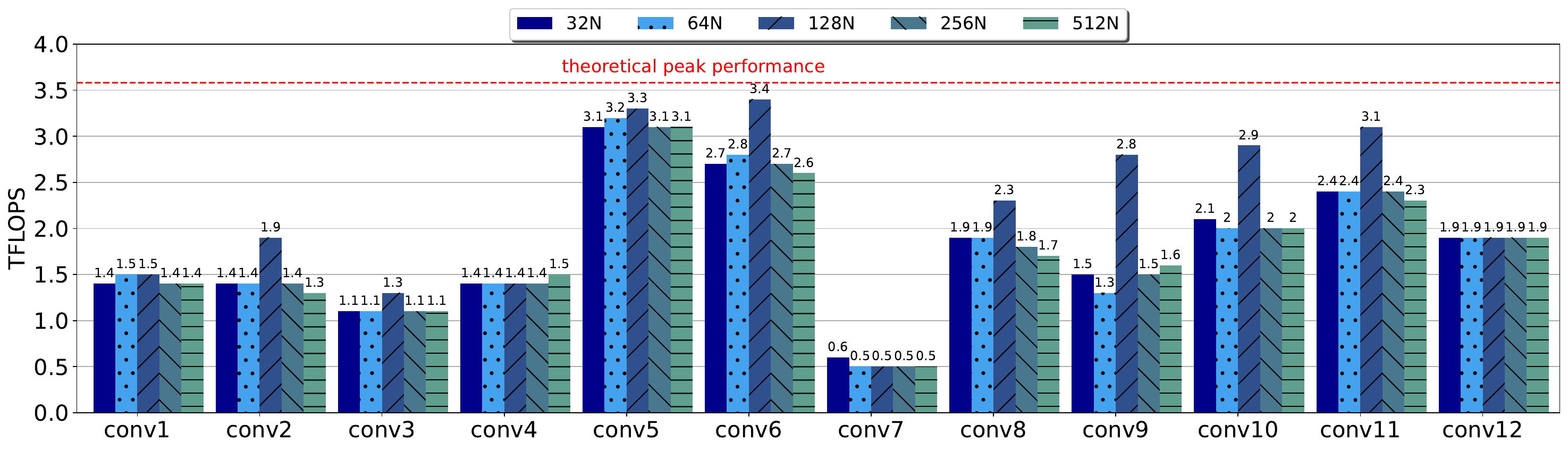}
	\caption{\small The performance of the direct convolution in different batch sizes with the NHWC layout}
	\label{fig:direct_nhwc_batch_scaling}
\end{figure*}

\begin{figure*}[ht]
	\centering
	\includegraphics[width=.9\textwidth]{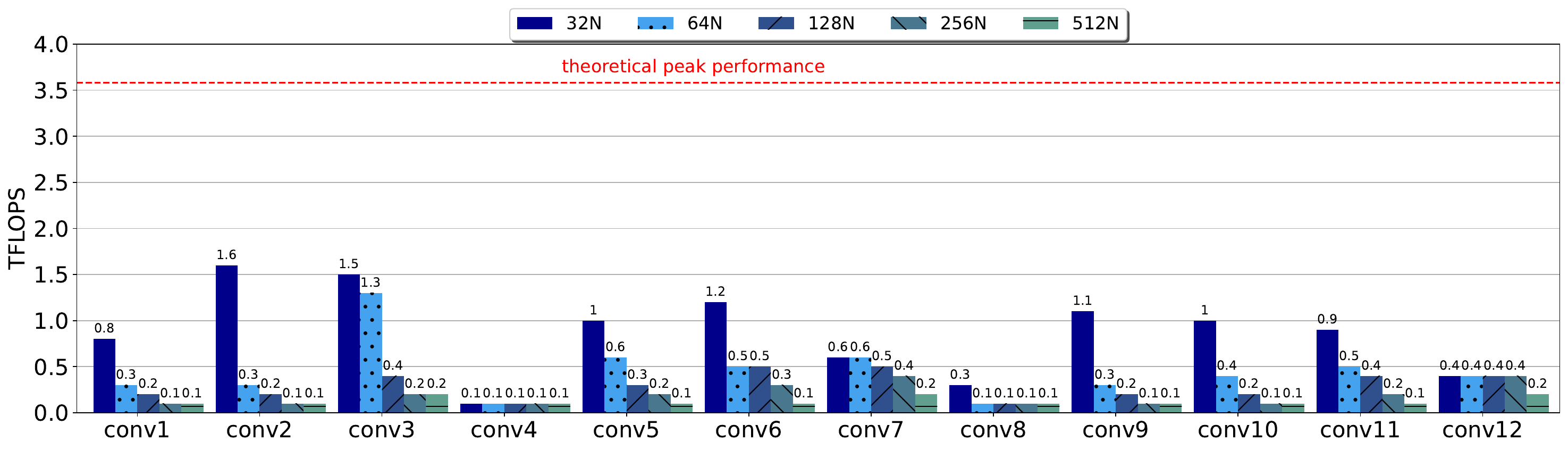}
	\caption{\small The performance of the im2win convolution in different batch sizes with the CHWN layout}
	\label{fig:im2win_chwn_batch_scaling}
\end{figure*}

\begin{figure*}[ht]
	\centering
	\includegraphics[width=.9\textwidth]{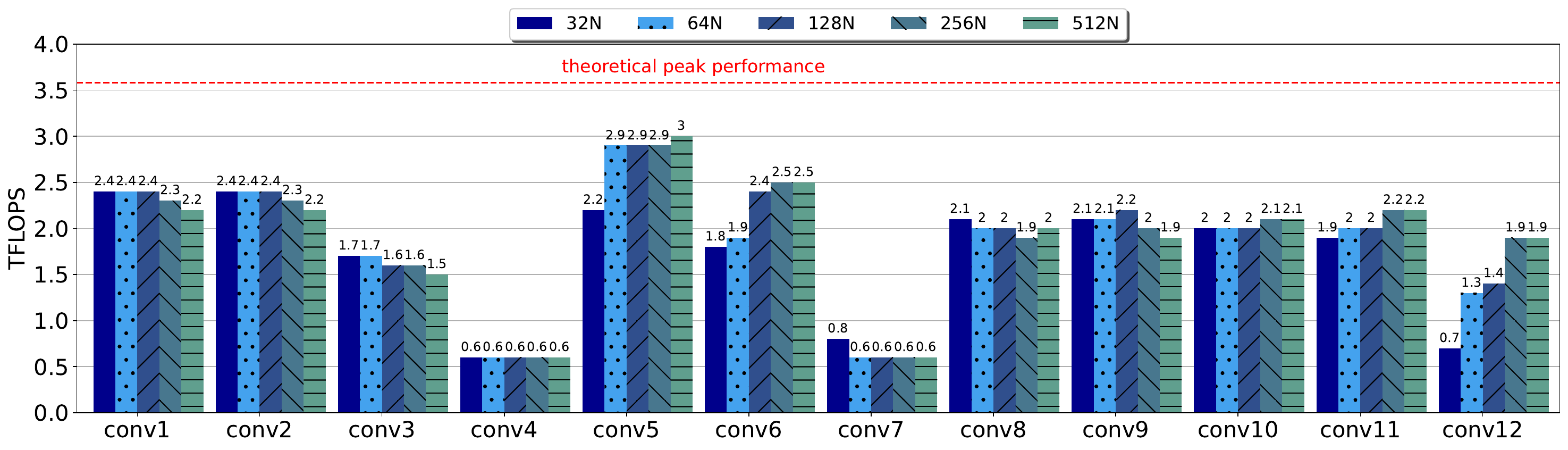}
	\caption{\small The performance of the im2win convolution in different batch sizes with the CHWN8 layout}
	\label{fig:im2win_chwn8_batch_scaling}
\end{figure*}

\begin{figure*}[ht]
	\centering
	\includegraphics[width=.9\textwidth]{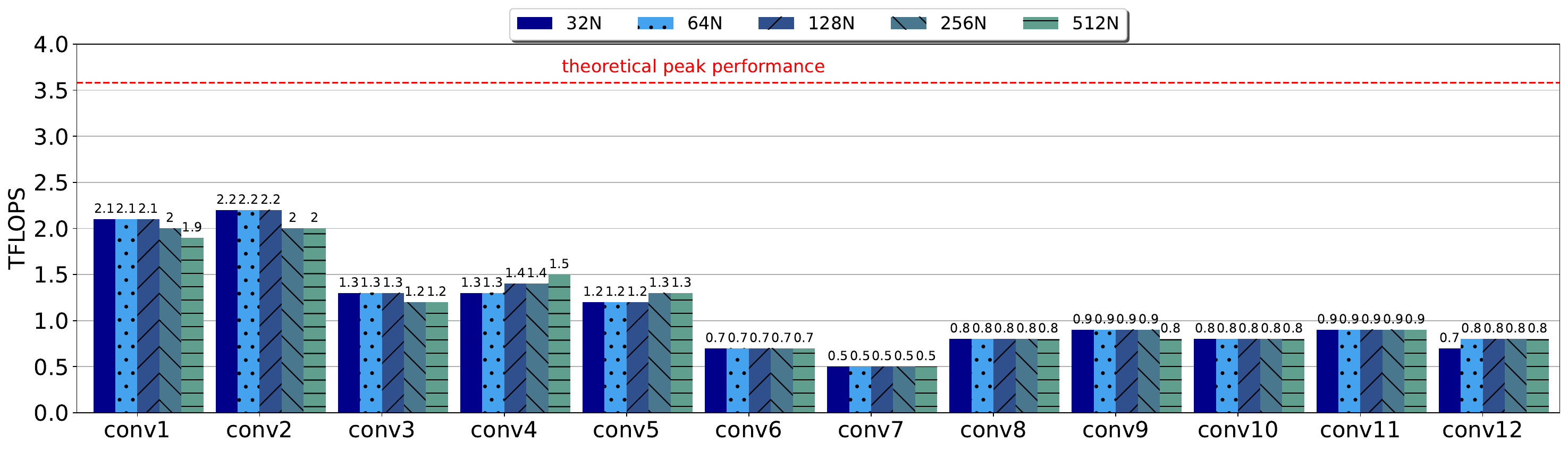}
	\caption{\small The performance of the im2win convolution in different batch sizes with the NCHW layout}
	\label{fig:im2win_nchw_batch_scaling}
\end{figure*}

\begin{figure*}[ht]
	\centering
	\includegraphics[width=.9\textwidth]{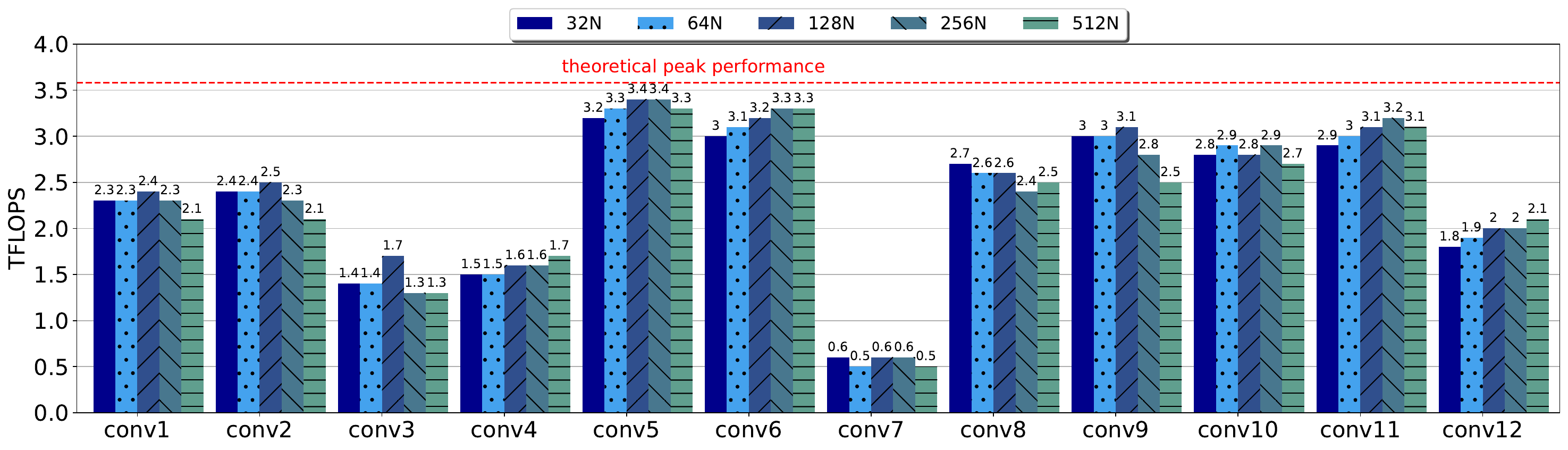}
	\caption{\small The performance of the im2win convolution in different batch sizes with the NHWC layout}
	\label{fig:im2win_nhwc_batch_scaling}
\end{figure*}

}

\end{document}